\documentclass[letterpaper, 10 pt, conference]{ieeeconf}  

\IEEEoverridecommandlockouts                              

\usepackage[backend=bibtex,
            hyperref=false,
            url=false,
            isbn=false,
            doi=false,
            backref=false,
            style=numeric-comp,
            sortcites,
            block=none]{biblatex}

\renewcommand{\bibfont}{\small}
\usepackage{courier}
\usepackage{graphicx}
\usepackage{ifthen}
\usepackage{float}
\usepackage{fancyvrb}
\usepackage{soul}
\usepackage{tabularx}
\usepackage{amsfonts} 
\usepackage{xcolor}
\usepackage{ctable} 
\usepackage{caption}
\usepackage{subcaption}
\usepackage{multirow}
\usepackage{hyperref}

\let\labelindent\relax  
\usepackage{enumitem}

\newcommand{\learnskill}{\texttt{learn\_skill}}
\newcommand{\pickskill}{\texttt{pick}}
\newcommand{\placeskill}{\texttt{place}}

\newcommand{\myitem}{\vspace{-1.5pt}\item}

\newboolean{draft-mode}
\setboolean{draft-mode}{false}

\DeclareRobustCommand{\asnote}[1]{\ifthenelse{\boolean{draft-mode}}{\textcolor{cyan}{\textbf{Anthony:} #1}}{}}

\DeclareRobustCommand{\pulkit}[1]{\ifthenelse{\boolean{draft-mode}}{\textcolor{orange}{\textbf{Pulkit:} #1}}{}}

\title{\LARGE \textbf{Lifelong Robot Learning with Human Assisted Language Planners}}

\author{
\authorblockA{Meenal Parakh$^{*,\alpha,\gamma}$, Alisha Fong$^{*,\alpha,\gamma}$, Anthony Simeonov$^{\alpha,\gamma}$, Tao Chen$^{\alpha,\gamma}$, Abhishek Gupta$^{\alpha,\beta,\gamma}$,  Pulkit Agrawal$^{\alpha,\gamma}$ \\ 
$^{\alpha}$Improbable AI Lab $\quad$ 
$^{\beta}$University of Washington $\quad$
$^{\gamma}$Massachusetts Institute of Technology \\
$^*$Authors contributed equally}
}

\begin{document}

\maketitle
\thispagestyle{empty}
\pagestyle{empty}

\begin{abstract}
Large Language Models (LLMs) have been shown to act like planners that can decompose high-level instructions into a sequence of executable instructions. However, current LLM-based planners are only able to operate with a fixed set of skills. We overcome this critical limitation and present a method for using LLM-based planners to query new skills and teach robots these skills in a data and time-efficient manner for rigid object manipulation. Our system can re-use newly acquired skills for future tasks, demonstrating the potential of open world and lifelong learning. We evaluate the proposed framework on multiple tasks in simulation and the real world. Videos are available at: 
\url{https://sites.google.com/mit.edu/halp-robot-learning}

\end{abstract}

\section{Introduction}
\label{sec:introduction}

    A dream shared by many roboticists is to instruct robots using simple language commands such as ``clean up the sink.''
    Large language models (LLMs) can support this dream by decomposing an abstract task into a sequence of executable actions or ``skills''~\cite{huang2022language}. Several LLM-based works use a \emph{fixed} set of skills (i.e., \emph{skill library})  for planning~\cite{saycan2022arxiv, huang2022inner}. %
    However, the available skills may not suffice in certain task scenarios.
    For instance, given the task, ``clean up the sink'', an LLM may plan a sequence of picks and places that move all the dishes to a dishrack. 
    Suppose one cup contains water which must be emptied before the robot puts it away. Without access to an ``empty cup'' skill, the system is fundamentally incapable of achieving this task variation.
    On detecting failure, LLM planners may attempt to expand their abilities -- the system could \emph{request} a new skill for ``pouring'' if it detects water in the cup. However, unless the robot can also \emph{execute} new skills, the problem remains unsolved. 
    
    Based on the tasks and scenarios the robot encounters, the planner must have the capacity to request and acquire \emph{new skills}. 
    Further, such skill acquisition ought to be \emph{quick} -- a system that requires days, weeks, or months to acquire the new skill is of little utility.
    Concurrent to our work, the ability of an LLM-based planner to acquire new skills has been demonstrated in the virtual domain of Minecraft~\cite{wang2023voyager}. However, in virtual domains, new skills can be simply represented as code that can execute high-level and abstract actions. In contrast, learning a new skill for a robot also involves finding low-level actions that can affect the physical world. To the best of our knowledge, the ability to add skills to a skill library in a time and data-efficient manner and utilize them for future tasks, especially in the context of LLM-based planners, has not been demonstrated.

    Existing LLM-based robotic systems struggle with online skill acquisition because common mechanisms for learning skills (e.g., end-to-end behavior cloning or reinforcement learning) typically require a large amount of data and/or training time.
    Some methods are able to acquire new skills in a more data-efficient manner in limited scenarios such as in-plane manipulation (e.g., TransporterNets~\cite{zeng2020transporter}), but these skills are insufficient for 6-DoF actions (e.g., ``grasp the mug from the side'', ``hang the mug on a rack" or ``stack a book in a bookshelf"). Another body of work such as in few-shot imitation learning can efficiently solve new instances from a task family but requires large amounts of pre-training data~\cite{duan2017one,pathak2018zero} which is seldom available for new skills. 
    We first present a method that allows LLMs to request new skills to complete the given task. Second, we propose to use Neural Descriptor Fields (NDFs)~\cite{simeonov2022neural} to realize these new skills. We choose NDFs as they require only 5-10 demonstrations to perform rigid body manipulation in the full space of 3D translations and rotations. 
    
    Our system works by prompting an LLM with a textual scene description obtained by a perception system, a library of skills expressed as Python functions, and a natural language task specification. With this information, the LLM plans and produces a sequence of skills (in the form of code) that achieves the task. Along with the skills in the skill library, we also provide the LLM with a special function for requesting a new skill to be added to the library. When the LLM plans call this \learnskill{} function, it returns a new skill name and a docstring description of the skill. However, such a skill is abstract and is not mapped to actions.  
    NDFs allow the user to quickly realize this new skill by providing a few demonstrations, after which the skill is added to the skill library so that it can be re-used on future tasks.
    In summary, this work demonstrates a proof-of-concept implementation of an LM-powered robotic planning agent that can interactively grow its skill library based on the needs of the task. We show an instance of such a system using NDFs and perform experiments that highlight the abilities of our system.  
    

\section{Related Work}
\label{sec:related}

\paragraph{LLMs as Zero-Shot Planners} 
Prior work that uses large language models (LLMs) as planners include SayCan \cite{saycan2022arxiv}, InnerMonologue \cite{huang2022inner}, NLMap-SayCan \cite{chen2022nlmapsaycan} and Socratic Models \cite{zeng2022socratic}. These methods make significant contributions: \cite{saycan2022arxiv} and \cite{zeng2022socratic} using LLMs as planners; \cite{huang2022inner} emphasizes the importance of feedback; and \cite{chen2022nlmapsaycan} improves upon \cite{saycan2022arxiv} by introducing the ability of open-vocabulary detection for grounding using CLIP and ViLD features \cite{radford2021learning} \cite{Gu2021OpenvocabularyOD}. The planners in these methods either generate the plan in textual format or choose the next step based on a given set of actions described through text.  Another set of methods \cite{codeaspolicies2022} \cite{vemprala2023chatgpt} \cite{singh2022progprompt}  \cite{lin2023text2motion} using LLM as planners chose to output the plans directly using a Python or symbolic API, given the function documentation and sufficiently expressive function names. 

\paragraph{End-to-End Language Conditioned Manipulation} 
Another class of methods processes inputs from different modalities such as visual, textual, and sound, and train an LLM to use these inputs to output robot actions end-to-end (e.g., CLIPort \cite{cliport}, Interactive Language \cite{lynch2022interactive}, RT1 \cite{brohan2022rt1}, PerAct \cite{pmlr-v205-shridhar23a} and VIMA \cite{jiang2022vima}). 
Another end-to-end approach is Palm-e~\cite{driess2023palme} that generates textual steps as output, and are assumed to map to a small set of low level policies. One main advantage these offer is more faithful LLM grounding, in contrast to modular approaches that only list the objects in the scene and sometimes fail due to partial scene descriptions. However, they each suffer from requiring a large amount of data for training or fine-tuning. Such large data requirements also make it difficult to achieve generalization. Finally, many of these works are limited to performing 3-DoF (top-down) manipulation actions. 

\paragraph{Low-Level Robot Primitives.} 
The modular approaches \cite{saycan2022arxiv} \cite{huang2022inner} \cite{chen2022nlmapsaycan} \cite{zeng2022socratic} \cite{codeaspolicies2022} \cite{vemprala2023chatgpt} use a predefined set of primitive skills, often hardcoded or learned from behavior cloning. These low-level primitives can also be learned through methods such as \cite{jang2021bc}, \cite{florence2022implicit}, \cite{chi2023diffusionpolicy}, \cite{zeng2020transporter}. While these skills can be composed to perform a wide range of actions, many times a required skill cannot be composed from the primitive set and adding a new primitive may require careful engineering, or large number of demonstrations. Thus, we employ \cite{simeonov2022neural} to incorporate new skills at runtime using only a few demonstrations, with the only drawback of limiting the skills to known object categories.


\section{Method}
\label{sec:method}

\begin{figure*}
    \centering
    \includegraphics[width=\linewidth]{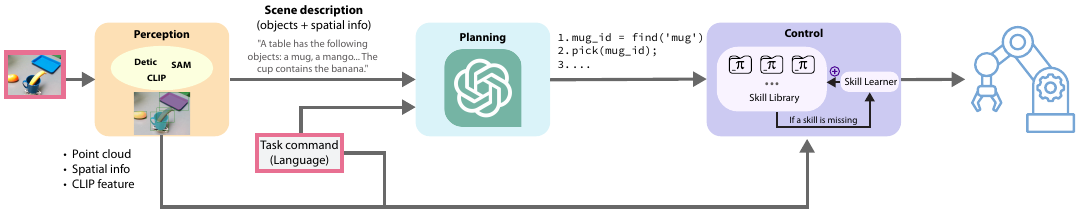}
    \caption{\footnotesize 
    Our system consists of three modules: \textit{perception}, \textit{planning}, and \textit{control}. The \textit{perception} module processes 
 RGB-D images and outputs a textual scene description that identifies objects and their spatial relationships. The \textit{planning} module uses GPT-4 to plan a sequence of steps based on the available skills and the task command. We added a \texttt{learn\_skill(skill\_name)} function to the planner so that it can plan to learn a new skill if such learning is necessary for completing the task. Finally, the \textit{control} module executes the planned steps using the available skills or starts learning a new skill. }
    \label{fig:system-pipeline}
\end{figure*}

In the spirit of prior work on performing long-horizon tasks wherein a high-level planning algorithm chains together different low-level skills~\cite{garrett2021integrated,leonard2007team,montemerlo2008junior,zeng2022socratic}, our system has explicit modules for perception, planning, and control (Fig.~\ref{fig:system-pipeline}). 
The modularity of our system allows us to take advantage of state-of-the-art (SOTA) models like SAM~\cite{kirillov2023segment} for segmentation and GPT-4~\cite{openai2023gpt} for planning skill sequences.
At a high level, our perception module describes the scene from RGB and depth observations, generating a language-based scene description containing information about the objects in the scene and the spatial relationship between them. Given the scene description and a library of skills, the planning module plans a sequence of steps to solve the task based on the scene description and task requirements. 
The skill sequence corresponds to a set of executable behaviors on the robot.

In contrast to previous work that uses LLMs in robotics, our planning module can request to learn a new skill when it determines that the existing skills are insufficient, and a data-efficient skill learning method can be used to extend the skill library with this new executable behavior.  
With an expanded skill library, the planner can utilize both the original primitive skills and the newly learned skills when completing subsequent tasks. Thus, our approach endows the system with a form of continual learning. 
In the following subsections, we describe each module in detail.

\subsection{Perception}
\begin{figure}[!t]
  \centering
    \includegraphics[width=\linewidth]{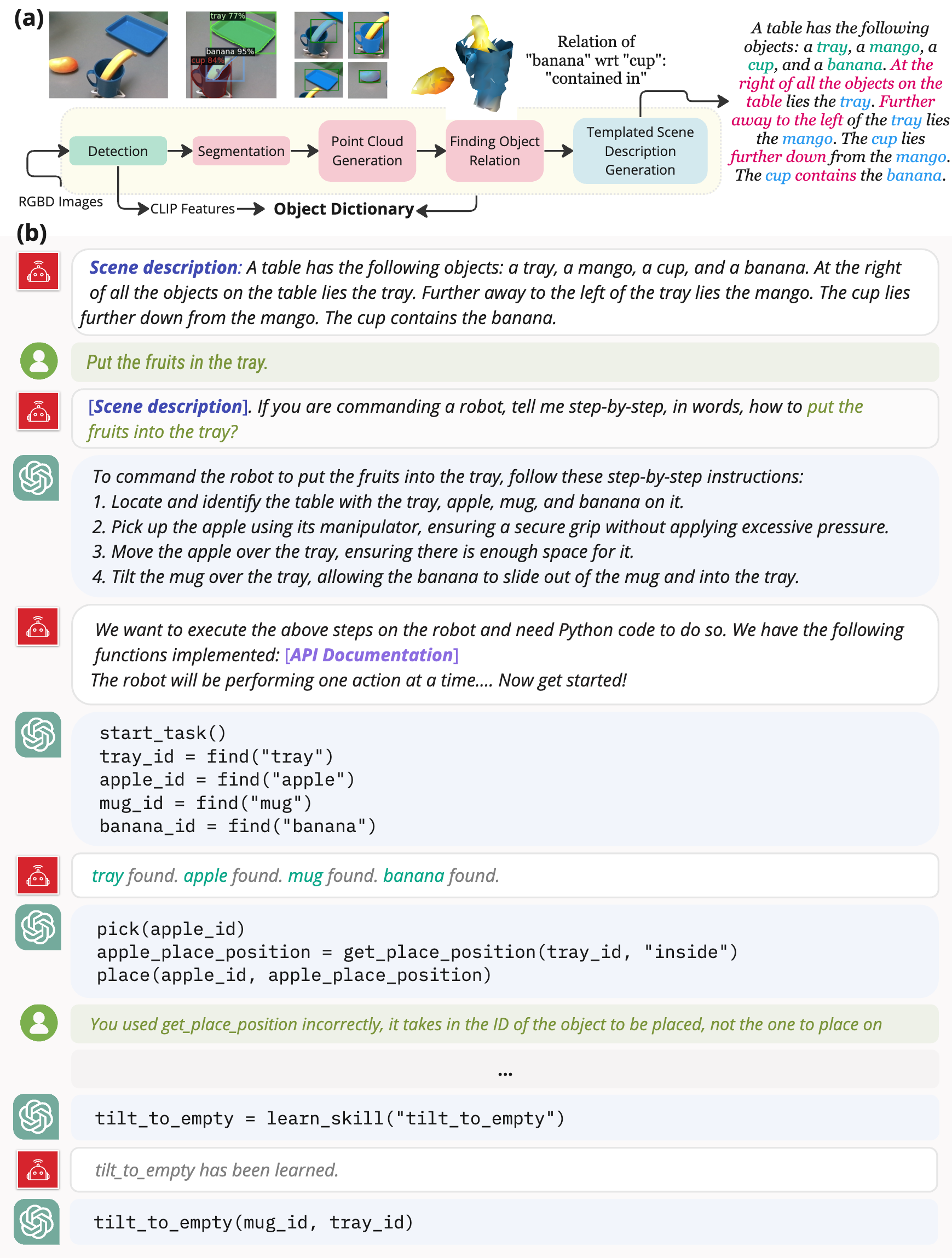}
    \caption{\footnotesize (a) From RGBD images, our perception module obtains information about the objects and their relations, creates an object information dictionary, and generates a scene description (\color{teal}{detection}, \color{cyan}{object pairs} \color{black} corresponding to given \color{magenta}{object relations}, \color{black} and the template is in black). (b) An example showing the interaction between the robot, the user, and the planner.}
    \vspace{-10pt}
    \label{fig:perception_planning}
\end{figure}

The perception module (Fig.~\ref{fig:perception_planning}a) processes RGBD images to obtain and store information about the scene objects.
First, the module identifies objects using an open-vocabulary object detector \cite{zhou2022detecting}. 
We also perform segmentation to obtain object masks using SAM \cite{kirillov2023segment} and combine them with the depth images to obtain object point clouds. 
In addition to object labels and segmentation masks, the planner may require additional information about the spatial arrangement of the scene. 
For example, if a robot needs to empty a mug, it first needs to know whether there is an object \emph{in} the mug, and only execute the skill of emptying it if there is. 
We generate spatially-grounded scene descriptions automatically by computing positional relationships between objects using the object point clouds. A scene description that is not spatially grounded only describes the objects present in the scene, without specifying the spatial relationship between them.
Lastly, to enable open-vocabulary language commands that target specific object instances, we extract CLIP embeddings of each segmented object in the scene. In this way, given a scene with multiple mugs, if the task is to ``pick up the red mug," we are able to identify the object that corresponds to the description of a ``red mug" (additional examples in Appendix). 
Overall, our perception components output segmented object point clouds with associated detection labels, inter-object relations, and CLIP embeddings. 

\paragraph{Spatially-grounded Textual Scene Description}
To inform the planner about the environment state, we format the perception outputs into a language-based scene description with information about the scene objects and their inter-object relations.
This involves constructing a string with the names of the objects along with the relations that hold between them. The description is akin to a textual description of ``scene graph''.
Please see Appendix for further details. Note that the particular method of describing the scene is not critical to our work and in the future vision-language models capable of describing objects and the relationship between them can replace this system. 

\subsection{Planning and Control}
\label{sec:lmp_api}

Given the language command and the textual scene description from the perception system, GPT-4 is used to plan a sequence of the steps to be executed. The inputs and outputs of the LLM are structured as follows: 

\paragraph{Skill Definitions via Code API}
One way to design a planner is to output a plan in natural language. However, a more machine-friendly alternative is to have the planner output programming code~\cite{codeaspolicies2022, singh2022progprompt}. Having an LLM planner directly produce code avoids the need to map a textual plan to a robot-executable plan. In addition, communicating with LLM in a programming language allows a human to give prompts in the form of comments, docstrings, and usage examples, which helps the planner understand how each skill operates. 
To take advantage of these benefits, we define each skill as a Python function that takes input arguments such as object identifiers and environment locations. We provide the planner with a description and set of input/output examples for each function. 
The code API is initialized with a skill library $\mathcal{S}_{0}$ containing five primary functions: \texttt{find}, \texttt{pick}, \texttt{get\_place\_position}, \texttt{place}, and \learnskill{}: 

{
\begin{itemize}[leftmargin=*]
    \myitem \texttt{find(object\_label=None, visual\_description=None, location\_description=None)}: 
    searches with the perception system for an object based on category, visual property, or location. Returns an \texttt{object-id}.
    \myitem \texttt{pick(object\_id)}: uses Contact-GraspNet \cite{sundermeyer2021contactgraspnet} to find a 6-DoF grasp for the object point cloud associated with the \texttt{object\_id} and executes the grasp.
    \myitem \texttt{get\_place\_position(object\_id, reference\_id, relation)}: for the object given by \texttt{object\_id}, returns the $(x, y, z)$ location determined by the text description \texttt{relation} relative to \texttt{reference\_id} .
    \myitem \texttt{place(object\_id, place\_position)}: places the object at the $(x, y, z)$ value given in \texttt{place\_position}.
    \myitem \learnskill{}\texttt{(skill\_name)}: returns a new executable skill function and a docstring describing the skill behavior.
\end{itemize}
}

The above API functions also output a signal indicating whether or not the function executes properly (i.e., to catch and correct runtime errors due to syntax mistakes).
If new skills are learned (discussed in Sec. \ref{sec:learning-new-skills}), the library is updated $\mathcal{S}_{i} = \mathcal{S}_{i-1} \cup \{\pi_{i}\}$ where $\pi_{i}$ denotes the new skill. %

\paragraph{Full Planner Input/Output and Skill Execution}
The planner is prompted to produce the plan in two steps. First, given the scene and task description, the planner generates a sequence of steps described in natural language. Next, the planner is provided with the code API of skills as discussed above and tasked to write code for executing the task using the given skills. For example, if the first step in the plan is to ``find'' a mug with the \texttt{find} function, the planner may output \texttt{object\_id = find("mug")}. Since our system uses a LLM planner, the human user can interact with the planner at either stage of the planning to further refine the plan or correct mistakes. An example of the interaction between the user, planner, and robot is shown in Fig.~\ref{fig:perception_planning}.
We qualitatively observe this two-step process helps the model generate higher-quality plans, as compared to producing the full plan directly. The two-step breakdown potentially helps in the same way ``chain-of-thought'' prompting has helped LLM find better responses~\cite{wei2022chain}. 

The code returned by the LLM is executed using the \texttt{exec} construct in Python. For skills involving robot actions, the skill function calls a combination of inverse kinematics (IK), motion planning, and trajectory following using a joint-level PD controller.

\subsection{Learning New Skills and Expanding the Skill Library}
\label{sec:learning-new-skills}
\label{sec:control}
\paragraph{Requesting New Abilities with \learnskill{} function}
The code API for the \learnskill{}, contains a docstring detailing the role of the function and also includes a few examples of the desired output of using the \learnskill{} function. The reason for providing examples is to exploit the in-context learning ability of LLMs -- these examples help the LLM figure out how to use the \learnskill{} function.  More details are in the Appendix. 
The \learnskill{}(\texttt{skill\_name}) returns the handle to a new executable skill function along with a docstring that describes the behavior of the function. 
The function is parameterized by either one or two \texttt{object\_id}s - one for specifying which object \texttt{skill\_name} acts upon, and another for specifying a reference object for relational skills (e.g., \texttt{pick(bottle\_id)} vs. \texttt{insert(peg\_id, hole\_id)}). The exact parameterization is decided by the LLM. 
When \learnskill{} is called, the returned function is added to the skill library so that the new skill can be reused in the future.

\paragraph{Data- and time-efficient skill grounding with NDFs} 
Our framework is agnostic to the specific method used to ground newly learned skills into actions. It can be end-to-end learning with reinforcement learning, or behavior cloning from demonstrations. In this work, we choose to use NDFs~\cite{simeonov2022neural} to learn new skills because it allows efficiently learning a skill from just a few ($\leq$10) demonstrations.
NDFs also facilitate a degree of category-level generalization across novel object instances, as well as generalization to novel object poses due to built-in rotation equivariance. 
More information on NDFs can be found in \cite{simeonov2022neural, rndfs}.

\paragraph{Learning from Feedback}
\label{subsec:learning_feedback}
If we specify a task the system cannot solve using the available skills (such as ``pick up the mug by the handle'', when the available ``pick'' skill grasps the mug from the rim), we would expect the LLM to directly request a new skill with \learnskill{}. While this occurs the majority of the time (see Experiments Section), the planner sometimes directly attempts the task using a skill that does not satisfy the task requirements. 
In these cases, if a user provides the \emph{outcome} of a task attempt (e.g., ``the mug was grasped by the rim''), the planner can use this information to register its usage of an incorrect skill and subsequently call \learnskill{} to expand its abilities. The system can then attempt the task with the newly learned skill. 

This highlights the need for feedback mechanisms that, in addition to detecting runtime errors, also inform the planner about the state of the environment after skill execution. To achieve this, we allow a human operator to manually but \textit{optionally} provide feedback before and after code execution. We allow the human to provide feedback after the execution of every step in the code. 
The combination of \textit{outcome} feedback from the user and the \textit{execution} feedback from the skill functions enables the system to detect failures, replan and if necessary expand its skillset using \learnskill{}. 

\iftrue{}
\paragraph{Continual Learning}
Learning new skills allows one to execute a task that was previously not possible. However, the full potential of learning new skills is realized when we allow the system to \emph{continually} acquire and \emph{re-use} skills to solve future tasks. This creates a system with ever-expanding capabilities.
There are many ways this can be achieved -- our implementation involves simply adding a new skill function expressed as a code API to the skill library, and using the updated library for future tasks.  
\fi

\section{Experiments}
\label{sec:result}

\paragraph{Environment Design and Setup}
We design our experiments to achieve three goals: (1) Show a proof-of-concept implementation of LLM-based task planning and execution with interactive skill learning in the real world, (2) Evaluate the abilities of current LLMs to appropriately request and re-use new skills based on the needs of different manipulation tasks, and (3) Compare the performance of the system when different components (such as object relations) are included vs. removed.  

In the real world, we tested our system on the Franka Panda robot with a Robotiq 2F-140 parallel jaw gripper. We used four calibrated RealSense cameras to obtain RGB-D images and point clouds.
We also evaluated the LLM planner in isolation with a set of manually crafted tasks, scene descriptions, and success criteria.
To perform additional system ablations, we evaluate our approach in simulation using PyBullet~\cite{coumans2016pybullet} and the AIRobot library \cite{airobot2019}. Our environment includes a tabletop-mounted Panda with the default gripper, and synthetic cameras for obtaining RGB-D images and segmentation masks. We use a combination of ShapeNet~\cite{shapenet2015} and manually-generated objects for experiments in simulation.

\begin{figure*}[t]
    \centering
    \includegraphics[width=\linewidth]{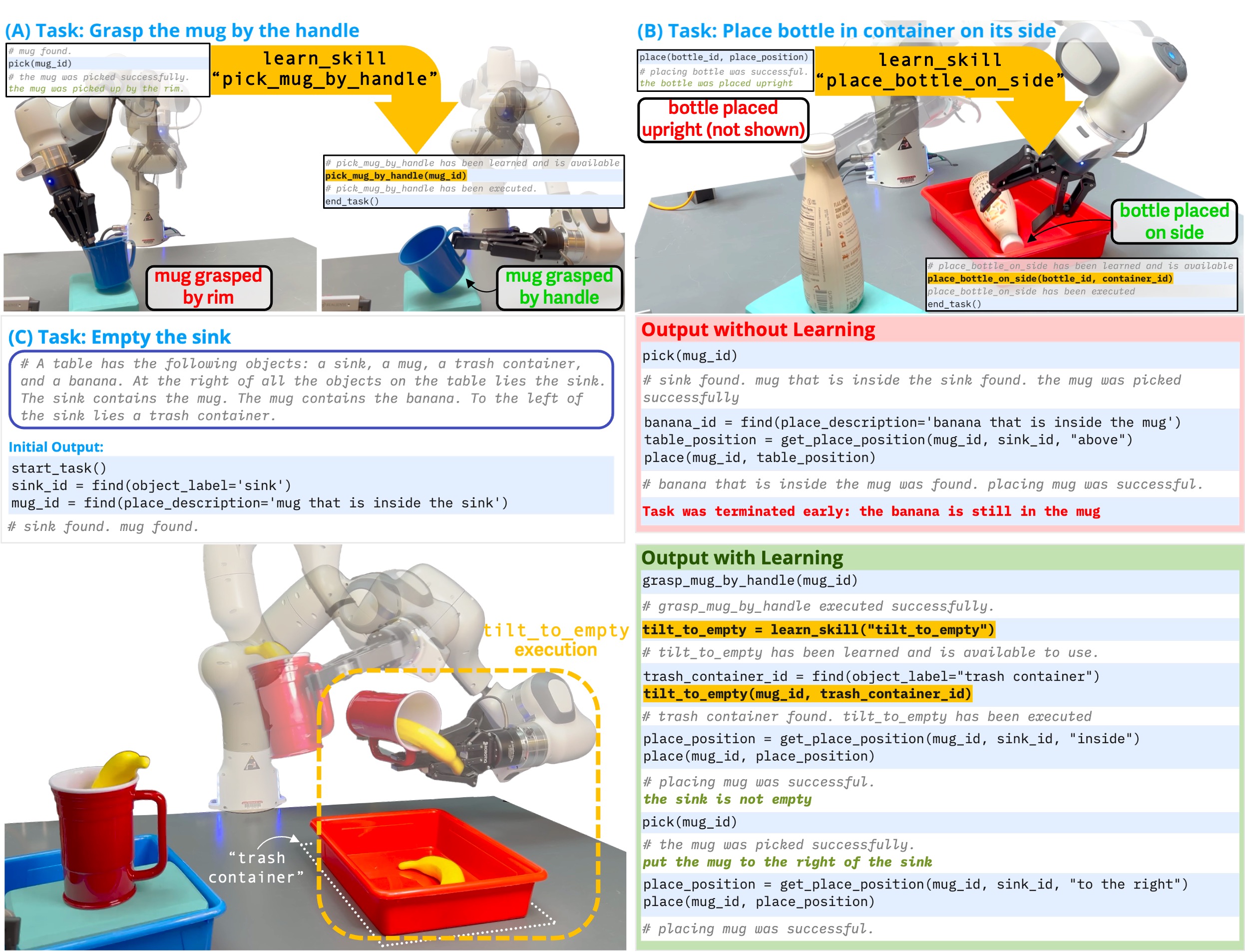}
    \caption{\small High-level plan and images for three tasks requiring a new skill: (A) Grasp mug by the handle, (B) Place bottle in container on its side, and (C) Empty the sink. The gray comments represent execution feedback while the green text is human feedback. When \learnskill{} is not available, the robot fails to complete the tasks. However, by learning new skills, the planner expands its abilities and satisfies each task requirement.}
    \vspace{-10pt}
    \label{fig:learn-skill-qual-results}
\end{figure*}

\subsection{Real-world tasks requiring \learnskill{}}
We first showcase the benefits of incorporating \learnskill{}. The system is deployed to perform three tasks in the real world: (1) grasping a mug by a specific part, such as the handle, (2) placing a bottle in a container that must fit on a small shelf, and (3) emptying a mug from a ``sink''.
Each task can be completed in multiple ways, some of which do not fulfill the full set of task requirements.
Our reference point for comparison is the overall system with no feedback mechanism and no \learnskill{} capability. This version directly attempts each task using the base set of primitive skills. 
Below, we discuss the differences between this baseline and the full version of our system.
The full set of planner inputs/outputs for these tasks can be found in the Appendix.

\subsubsection{Learning and requesting new pick and pick-place skills}
\textbf{Task 1: Grasp mug by handle}
Our warm-up task that highlights how learning new skills can benefit our system is to perform grasping by a specific part. In this case, we ask the system to ``grasp the mug by the handle'' (see Fig.~\ref{fig:learn-skill-qual-results}A). 
Without \learnskill{}, the planner directly calls \pickskill{} on the mug. This triggers a grasp detector~\cite{sundermeyer2021contactgraspnet} to output a set of grasps on the corresponding mug point cloud. 
Since most of these grasps are along the rim of the mug, the robot executes a grasp along the rim of the mug, and the task finishes.

If an incorrect skill is used, the human can prompt the system with feedback. 
By telling the system ``the mug was picked up by the rim'', the planner puts the mug back down and requests to learn a new \texttt{pick\_mug\_by\_handle} skill.
We teach this as a side-grasp at the handle using NDFs with five demonstrations.
After collecting the demos, we add  \texttt{pick\_mug\_by\_handle} to the skill library. Finally, the LLM directly calls \texttt{pick\_mug\_by\_handle} and finishes the task successfully. 

\textbf{Task 2: Place bottle in flat tray} Our next task is to place a bottle in a container that must eventually fit in a small shelf. 
Here, we prompt the system to ``place the bottle sideways in the container'' (see Fig.~\ref{fig:learn-skill-qual-results}B). 
When the pipeline runs using the base set of skills, the robot uses the only available ``place'' skill, which places the bottle upright in the tray.

Instead, when we provide the feedback ``the bottle was placed upright in the tray'', the LLM calls \learnskill{} to acquire a \texttt{place\_bottle\_sideways\_in\_tray} skill. 
This is implemented via NDFs as a side grasp on the bottle along with a reorientation and placement inside of the tray. 
Once this new skill has been added, the robot is able to successfully complete the task.

\subsubsection{Continual learning by re-using previously-learned skills}
\textbf{Task 3: Empty mug from sink} Finally, we prompt the system with the abstract objective of emptying a ``sink'' by removing a mug from the container and placing it on the table (see Fig.~\ref{fig:learn-skill-qual-results}C).
This task implicitly requires \emph{emptying} the mug before placing it.
We test the LLM's ability to satisfy this requirement by placing an additional small object (banana) inside the mug (ensuring the object is at least visible by the cameras, but difficult to pick up directly). 
The baseline system directly calls a combination of \pickskill~on the mug and \placeskill~to put the mug down on the table. 

However, with access to \learnskill{} and the dynamic skill library, the planner \emph{reuses} \texttt{pick\_mug\_by\_handle} learned in Task 1 and immediately requests to learn \texttt{tilt\_mug} so it can first move any objects in the mug to the trash container. 
We again use NDFs to teach \texttt{tilt\_mug}, which reorients the mug above the tray. 
After emptying, the system plans to place the mug back \emph{into} the sink. The user tells the system ``the sink is not empty, put the mug to the right of the sink''. Finally, the LLM re-plans with this feedback and achieves the final placement on the table. 

\subsection{LLM-only skill learning evaluation}
In this section, we examine the isolated ability of the LLM-planner to utilize the \learnskill{} function and to appropriately re-use and/or \emph{not} re-use newly-learned skills on subsequent runs. 
This enables further analysis of GPT-4's ability to interpret manipulation scenarios represented via textual scene descriptions and correctly use the available skills provided in the code API. 
For each task in the following subsections, we provide a manually-constructed scene description (that does not correspond to any particular real-world scene) along with a task prompt and the skill API.
We ask the planner to output code that completes the task using the API functions. The code output is manually evaluated as correct/incorrect by a human. 

\textbf{Requesting new skills when needed}
First, we study the ability to either (i) properly call \learnskill{} or (ii) properly \emph{not} call \learnskill{}, for a variety of tasks where either the base skill set is (ii) or is not (i) insufficient for the task, respectively.
We report the fraction of attempts that correctly use or ignore \learnskill{} in a scenario where human feedback is not provided. 
The results are shown in the top two sections of Table~\ref{tab:llm-only-eval}.
The 91\% success rate for using \learnskill{} without feedback indicates GPT-4 can be used for requesting an expanded skill set in a purely feed-forward fashion. 
Similarly, the LLM usually does not call \learnskill{} when it is not needed (87\% success). 
However, some performance gap remains in both settings.

\textbf{Re-using new skills with varying level-of-detail skill descriptions}
Next, we focus on the ability to properly re-use the previously-learned skills on subsequent runs, when they can either be applied or when they specifically should \emph{not} be applied (e.g., in scenarios where they are inappropriate or infeasible). 
We consider varying levels of detail in the description that accompanies the newly-learned skill as it is added to the code API. For instance, we can provide minimal information and only add the name of the new skill, or we can modify the return values of \learnskill{} so that the LLM writes its own docstring/function description to accompany the new skill when we add it to the API. 
The results are shown in the last two rows of Table~\ref{tab:llm-only-eval}.
The success rates indicate that the language model correctly uses the newly-learned skills with higher frequency when the skill descriptions also include docstrings. 
This makes intuitive sense, as it provides extra context for both the ability and applicability of the newly learned skill, which the LLM can attend to when generating the output code for executing the task (mimicking the chain-of-thought and ``let's think step-by-step'' improvements observed in prior work~\cite{wei2022chain, kojima2022large}). 

Despite the performance increase when describing newly-added skills in more detail, the LLM only achieves moderate overall performance (75\% success rate).
We observe this is due to a combination of sometimes using new skills when they should not be used (e.g., calling a \texttt{side\_pick\_bottle} skill even when the scene description says ``the bottle \emph{cannot} be reached from the side'') and re-learning the same skill multiple times (while occasionally calling it a very similar name) rather than directly utilizing the function that is already available in the API. 
We deem this as a somewhat negative result which points to potential gaps in such a method of LLM-based task planning. Namely, directly outputting a sequence of high-level skills (or exhaustively scoring them with a language model) does not allow more information about the operation of high-level skills (such as scenarios when they are or are \emph{not} applicable) to be provided or utilized during planning/reasoning.

\begin{table}
    \setlength{\tabcolsep}{1.5pt}
    \begin{center}
    \renewcommand{\arraystretch}{0.85}
    \begin{tabular}{@{}llc@{}}
        \toprule
        \textbf{Eval Metric} & \textbf{Variation} & \textbf{Success Rate} \\
        \midrule
        \textbf{Correct use of \learnskill{}} & -- & 0.91 \\
        \textbf{Correctly did \emph{not} use \learnskill{}} & -- & 0.87 \\
        \midrule
        \textbf{Correct re-use of new skill} & Name only & 0.50 \\
        \textbf{(varying skill description)} & Name + docstring & 0.75 \\
        \bottomrule
    \end{tabular}
\end{center}
\caption{Success rates for evaluations LLM-only \learnskill{} evaluation.}
\vspace{-15pt}
\label{tab:llm-only-eval}
\end{table}

\normalsize

\section{Limitations}
\label{sec:discussion}
While our system takes advantage of SOTA components, they sometimes fail and trigger compounding inaccuracies in the downstream pipeline. For example, the LLM heavily depends on an accurate description of the scene, which can sometimes contain erroneous detections and incorrect object relations.
We also leverage human feedback to obtain environment descriptions that inform task success and skill acquisition. Humans can provide accurate descriptions that inform when to learn new skills, but repeated user interaction makes the system less autonomous and slower to execute tasks. Leveraging learned success detectors would make the system more autonomous and self-sufficient. 
Similarly, human verification is typically needed to confirm the overall success or failure of a task, making it difficult to run system evaluation experiments at scale and limiting our evaluations primarily to qualitative demonstrations. 

\section{Conclusion}
\label{sec:conclusion}
This paper presents a modular system for achieving high-level tasks specified via natural language. Our framework can actively request and learn new manipulation capabilities, leading to an ever-expanding set of available skills to use during planning. We show how an LLM planner can use this  ability to adapt its skill set to the demands of real-life task scenarios via both feed-forward reasoning and environmental feedback. In conjunction with perceptual scene representations obtained from off-the-shelf components and a data-efficient method for learning 6-DoF manipulation skills, we provide an example of a complete system. Our results demonstrate how this combination of full-stack modularity, spatially-grounded scene description, and online learning enables a qualitatively improved ability to perform manipulation tasks specified at a high level. 


\iftrue{}
\section{Acknowledgement}
We thank the members of Improbable AI for their feedback on the project. 
This work is supported by Sony, Amazon Robotics Research Award, and MIT-IBM Watson AI Lab. Anthony Simeonov is supported in part by an NSF Graduate Research Fellowship. 

\vspace{10pt}
\subsection{Author Contributions}
\textbf{} \\
\textbf{Meenal Parakh} Co-led the project, developed the core LLM-planning framework and full-stack system, set up and ran experiments in simulation and the real world, and drafted the paper. \\
\textbf{Alisha Fong} Co-led the project, integrated NDF-based skill learning into the LLM-planning framework, set up and conducted experiments in the real world and simulation, helped evaluate the LLM in isolation, and drafted the paper. \\
\textbf{Anthony Simeonov} helped integrate NDF-based skills into the framework, supported real robot experiments and LLM-only evaluation, and helped revise the paper. \\
\textbf{Tao Chen} engaged in brainstorming and discussion about system implementation and experiment design, mentored Meenal Parakh, and helped draft the paper. \\
\textbf{Abhishek Gupta} was involved with technical discussions, advised Meenal Parakh, and helped with project brainstorming in the early phases. \\
\textbf{Pulkit Agrawal} advised the project and facilitated technical discussions throughout, helped refine the project focus on interactive skill learning with LLMs, and revised the paper. 

\fi



\renewcommand*{\bibfont}{\footnotesize}
\printbibliography 

\newpage

\renewcommand{\thesection}{A\arabic{section}}
\renewcommand{\thefigure}{A\arabic{figure}}
\setcounter{section}{0}
\setcounter{figure}{0}


\onecolumn

\def\centsect#1{\Large\centering#1}
\section*{\centsect{Appendix}}

\section{Additional Experiments}
\subsection{Benefiting from spatially-grounded scene description}
The next set of tasks we consider evaluates the benefit of providing a spatially-grounded scene description. We also consider basic tasks of picking objects and placing them in a described position, and the stacking task that involves a sequence of three pick-place actions. The main results are shown in Table~\ref{tab:task_lst}, where each row in the tasks column represent a set of tasks involving different object categories (for example, mugs, bowls, bottles) and receptacles (for example, containers, and baskets).  Each task in a row is performed for $\ge 10$ runs with varying object instances, and success or failure is assigned to each run, which is then used to find the success rate.

The first task category in scene description requires grasping a target object that lies beneath a second object. Without a relational scene description, the planner directly picks the target object. On the other hand, if the planner is informed that the second object is \emph{above} the target, it calls \pickskill{} and \placeskill{} on the second object before fetching the target object.

The second example showing the advantage of our scene description involves detecting task progress to minimize the number of actions that are used. For instance, to ``put all objects of a specified category into a basket'', the generated plan completed the task in fewer steps if the planner knows some objects are already in the tray. 

The final category of tasks calls for satisfying relational constraints with an \emph{indirectly specified} reference object. One example is ``place the apple in a tray \emph{without mugs}''. When provided with the scene description, the planner detects the tray without any mugs and uses it as the placing target. In contrast, without the scene description, the system fails to place in the correct tray.

\begin{figure*}[ht]
    \centering
    \includegraphics[width=12cm]{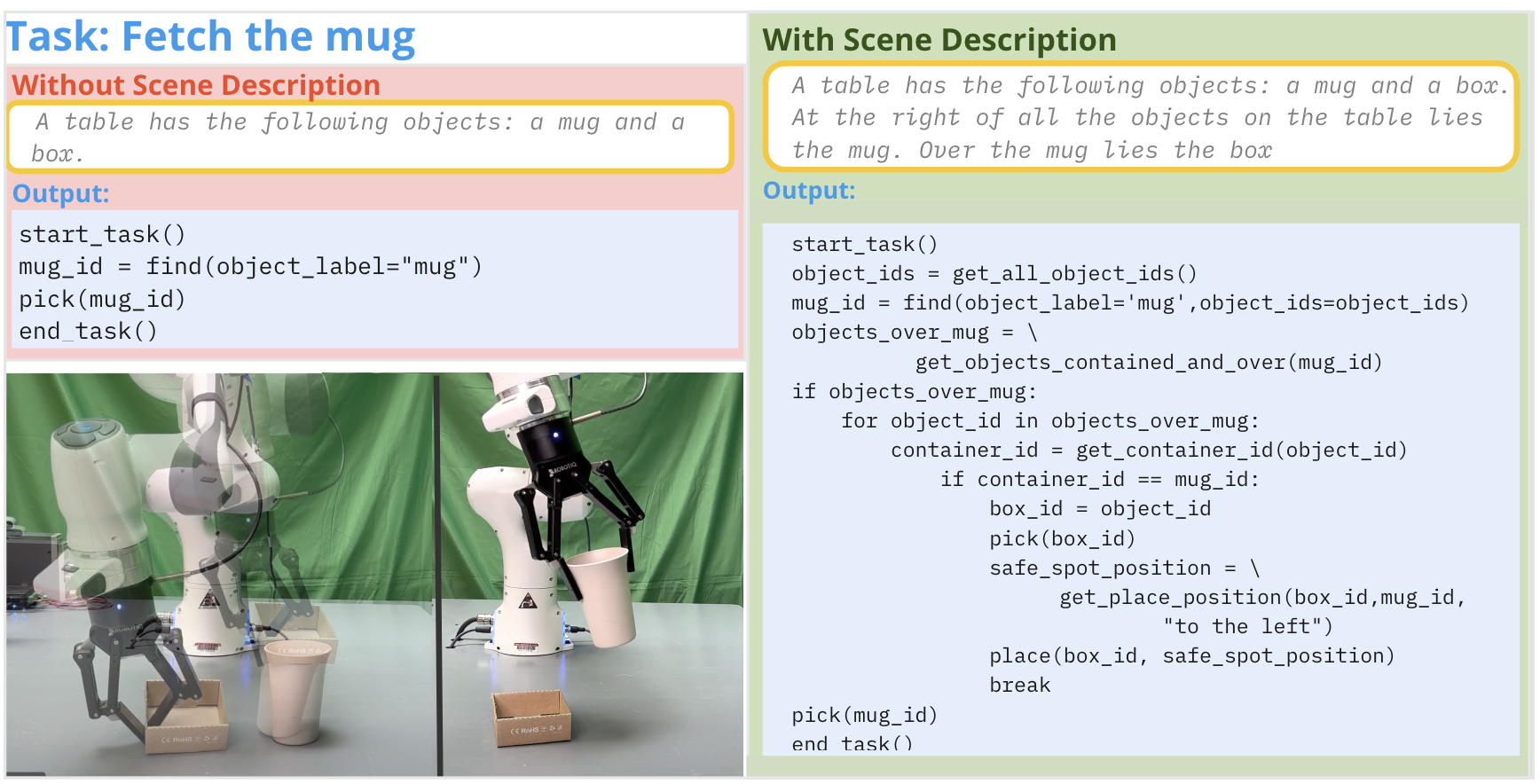}
    \caption{ Example plan for ``box over a mug" task, with and without spatially-grounded scene description information in the input. If the scene description lacks spatial information, the planner fails to communicate that the box must be removed before picking up the mug.}
    \label{fig:scene_desc_plan}
\end{figure*}

\begin{table*}[!h]
\vspace{-0.25cm}
\footnotesize
\begin{center}
\begin{tabular}{lll}
\specialrule{.2em}{.1em}{.1em} 
& Tasks & Success \\ 
\specialrule{.1em}{.05em}{.05em} 
\textbf{Basic Tasks} & Pick \texttt{object} & 91\%\\
& Place \texttt{object-1} to the left of \texttt{object-2} & 80\%\\  
& Place \texttt{object} into \texttt{receptacle} & 76\%\\
& Stack three bowls & 70\% \\

\specialrule{.05em}{1em}{0em} 
\textbf{Scene Description} & Fetch \texttt{object-1} (when \texttt{object-2} lies over it) & 79\% \\
 & Place all \texttt{object}s into the \texttt{receptacle} (partial progress) & 75\% \\
 & Place \texttt{object-1} into \texttt{receptacle} that has no \texttt{object-2} & 75\%\\

\specialrule{.1em}{.05em}{.05em} 

\end{tabular}
\end{center}
\caption{\label{tab:task_lst}Success rates for evaluations in simulation.}
\end{table*}

\normalsize

\vfill
\pagebreak

\section{Tunable System Parameters} \label{sec:tunable_params}

\footnotesize
\begin{table*}[h!]
\begin{center}
\begin{tabular}{| p{2.5cm} | p{1cm} | p{9cm}|} 

\specialrule{.15em}{.05em}{.05em} 
 Parameter & Value & Description \\ [0.5ex] 
\specialrule{.15em}{.05em}{.05em} 

Detection thd. & $0.3$ & Object detection threshold in Detic \\
\hline
Mask Erosion & $10$ & The number of pixels removed from the mask's boundary to lessen the impact of high depth noise near object's edges. \\
\hline
PCD Merging thd. & $0.03$ & To find correspondence between objects in images from different views. If the change in std deviation for two object point clouds in the two images is less than the threshold, they are merged to represent one object. \\
\hline
Overlap thd. & $20$ & In the top down projection of two objects convex hull intersection contains atleast the threshold points, then the objects belong to either the ``above" case or the ``contained" case. \\
\hline
Contained thd. & $0.1$ & If the percentage of an object's points that lie inside the convex hull of another object is less than this threshold, then the object is said to be ``contained-in" in the other object. \\
\hline
Grasp thd. & $0.8$ & Grasps (generated from \cite{sundermeyer2021contactgraspnet}) with predicted success value higher than the threshold are considered.\\
\hline
Place description thd. & $0.6$ & The threshold for \texttt{location-description} option in \texttt{find} primitive. Uses BERT \cite{Devlin2019BERTPO} for the finding the score.\\
\hline
Visual description thd. & $0.3$ & The threshold for \texttt{visual-description} option in \texttt{find} primitive. Uses CLIP \cite{radford2021learning} for the finding the score. \\
\hline
Object label thd. & $0.3$ & The threshold for \texttt{object-label} option in \texttt{find} primitive. Uses BERT \cite{Devlin2019BERTPO} for the finding the score.\\
 \hline  
\end{tabular}
\end{center}
\end{table*}

\normalsize

\vfill
\pagebreak

\section{Open Vocabulary Object Detection} 
\begin{figure}[H]
  \centering
  \includegraphics[width=0.7\linewidth]{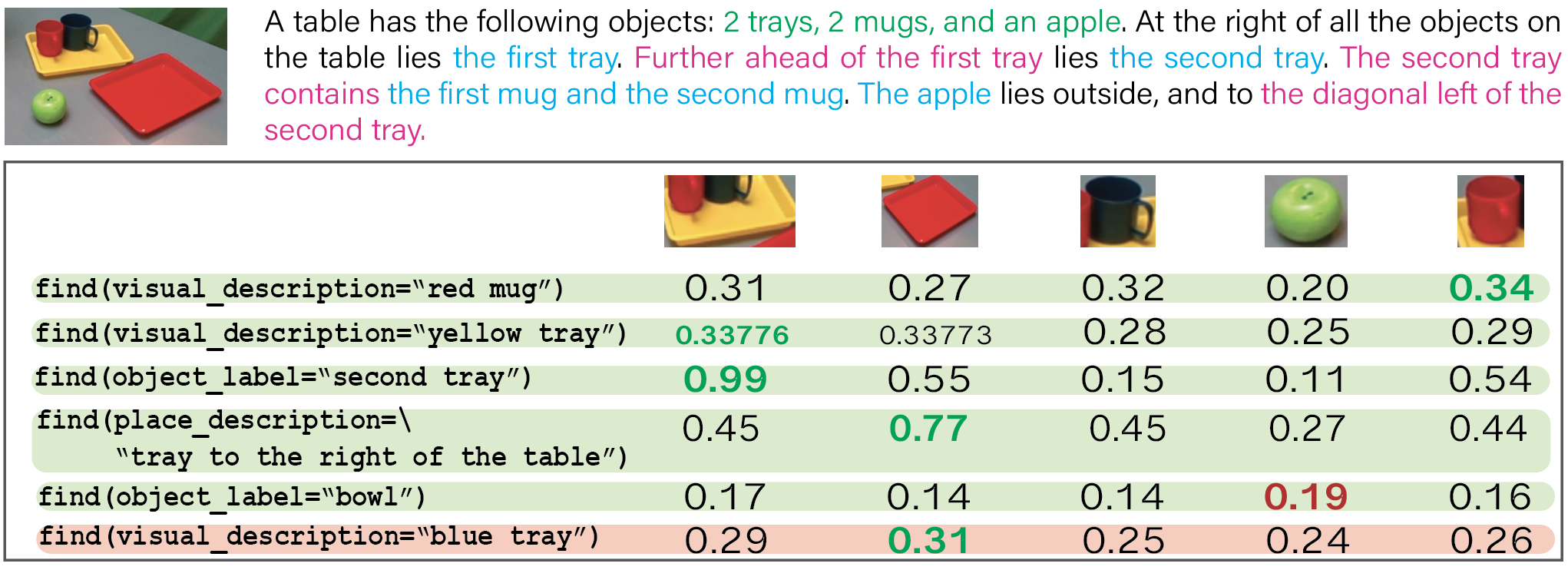}
    \caption{ Examples for \texttt{find} primitive in the API. Numbers in \color{teal} green \color{black} pass the threshold and the object is found. Number in \color{red} red \color{black} are below threshold and no object is found. The green highlighted are true positives, while the red highlighted is an example of a false positive.}
    \label{fig:find}
\end{figure}

The primitive function \texttt{find} performs the task of detecting any object in the scene, either through (a) an object label that comes from the scene description, (b) a visual description that describes an object's visual properties (identifiable through CLIP features), and (c) based on the object's location in the environment with respect to another object, as described through the scene description. Examples for different ways in which a call to \texttt{find} can be made is shown in Figure \ref{fig:find}
\label{sec:find_eg}

\section{Generating Scene Description} \label{sec:scene_desc}
\begin{figure}[H]
    \centering
    \setlength{\tabcolsep}{0.3pt}
    \begin{tabular}{cc}
    \begin{subfigure}[t]{0.4\linewidth}
        \centering 
        \includegraphics[width=0.99\linewidth]{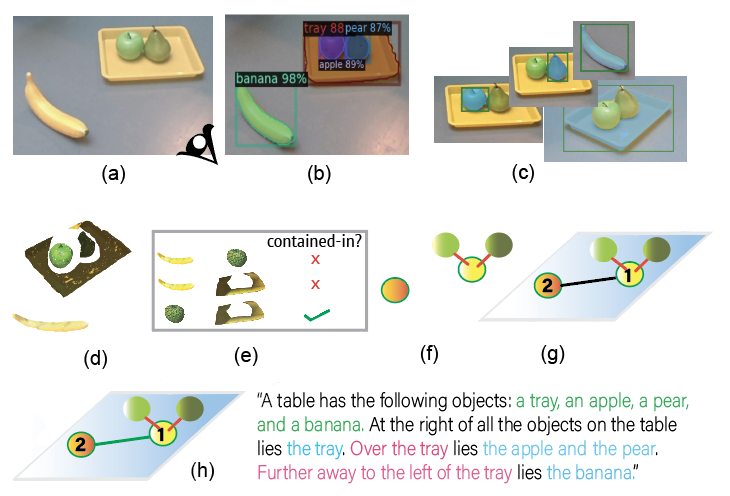}
        \caption{ Generating scene description involves steps (a) to (h): finding object point clouds, then finding spatial relations between pair of objects, followed by generating an object scene graph and defining relations over the edges and finally filling in the information in a template.}
        \label{fig:scene_desc}
    \end{subfigure} \hspace{30pt} \hfill
    \begin{subfigure}[t]{0.4\linewidth}
      \centering
      \includegraphics[width=0.99\linewidth]{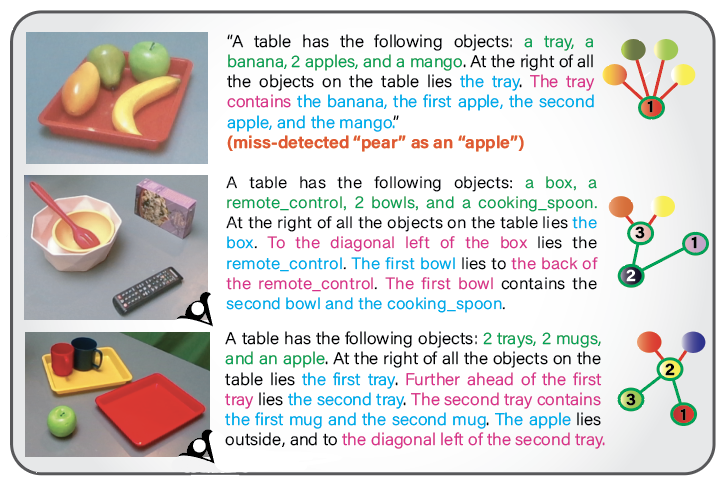}
        \caption{ Scene Description Examples: on the left are the example scene images, in which the user is standing on the right side of the image; at the center are the scene descriptions; and at the right side shows the object graphs that were traversed to produce the scene description.}
        \label{fig:scene_desc_exams}

    \end{subfigure}
    \end{tabular}

\end{figure}

Generating scene description involves the following main steps: finding object relations between pair of objects, generating an object scene graph and defining relations over the edges and finally filling in the information in a template. These steps from (a) to (h) are shown in Figure \ref{fig:scene_desc}. First, multiple RGBD images are obtained the camera sensors followed by detection (b) and segmentation (c) in those images. For each image, we obtain an object instance's partial point cloud which are merged with the partial point clouds from other camera images (d). Now each object's point cloud is compared with another object's; if they overlap and the convex hull of one covers some part of the object, then one is identified as containing the other object. Similar heuristic is used for identifying when an object lies above other object (e). Using these relations, objects are grouped together: a tray containing an apple will be treated as one vertex (f). Now all the grouped vertices are connected with one of their nearest vertex, and assigned one of the directional relations (left, right, front, down)(f). Next we traverse through the object graph and convert each edge, or relation into a sentence. While traversing the graph, on each grouped object node, the description of what objects are contained, or lies above within the node is also added to the description. Finally, we list all objects (and sum all objects of belonging to same category) and add the sentence to the beginning of the description, thus producing the final description (h). A few examples are shown in Figure \ref{fig:scene_desc_exams}.

\vfill
\pagebreak

\section{Pick and Place Primitives}
\begin{figure}[H]
    \centering
    \setlength{\tabcolsep}{0.3pt}
    \begin{tabular}{cc}
    \begin{subfigure}[t]{0.25\linewidth}
        \centering 
          \includegraphics[width=\linewidth]{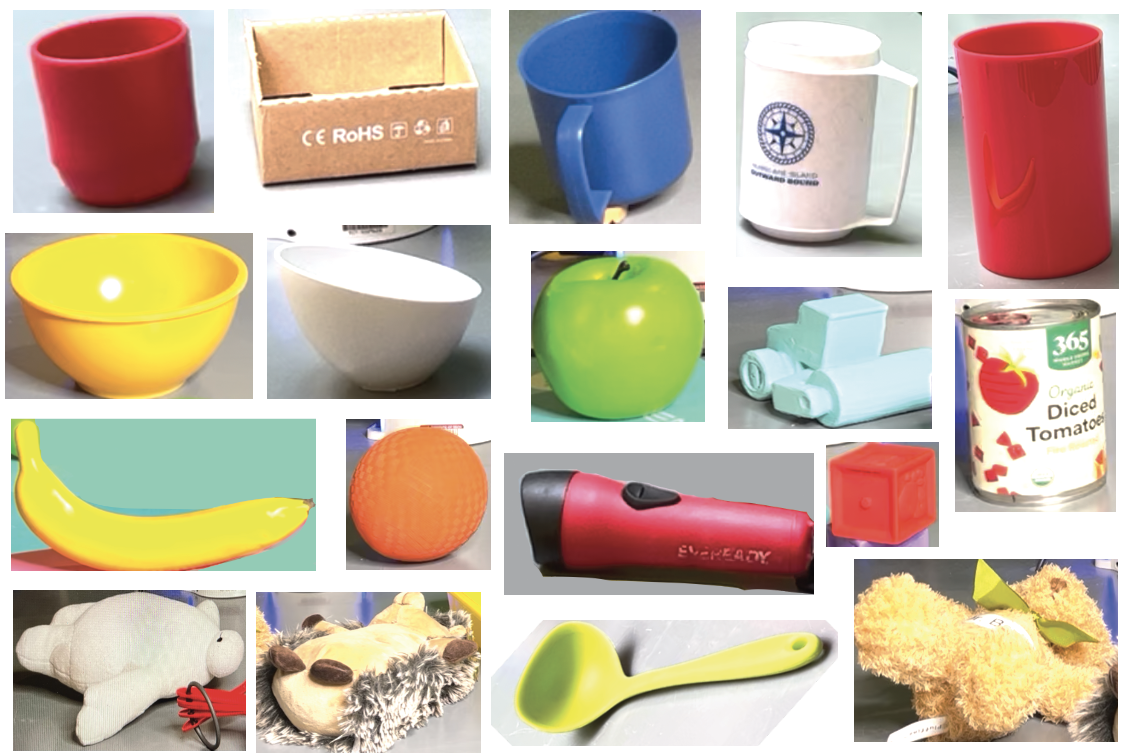}
            \caption{ Example real objects}
            \label{fig:real_objects}
    \end{subfigure} \hspace{10pt} \hfill
    \begin{subfigure}[t]{0.7\linewidth}
      \centering
          \includegraphics[width=\linewidth]{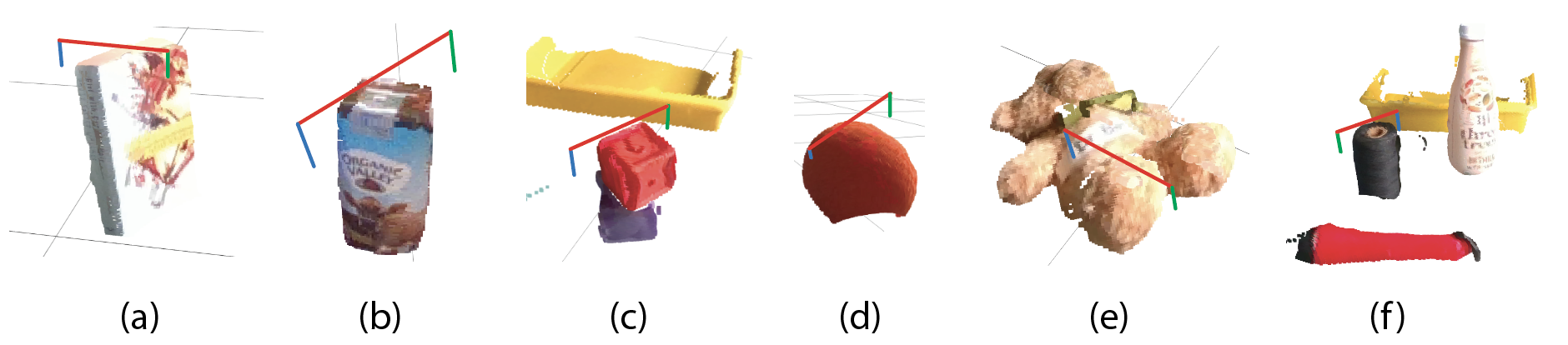}
            \caption{ Real World Grasp Examples. From left to right, showcasing the grasp for a book, a milk carton, a cube, a ball, a toy and a spool of thread.}
            \label{fig:real_objects_grasps}

    \end{subfigure}
    \end{tabular}

\end{figure}

Pick and Place primitives are main skills that the library has by default, prior to new skills being added. The pick function takes in an object id and access the object's point cloud, runs Contact Graspnet \cite{sundermeyer2021contactgraspnet} which is then executed. Some example objects we use for real world experiments are shown in Figure \ref{fig:real_objects} and some of the computed grasps are on those objects are shown in Figure \ref{fig:real_objects_grasps}.

\begin{figure}[H]
  \centering
  \includegraphics[width=0.7\linewidth]{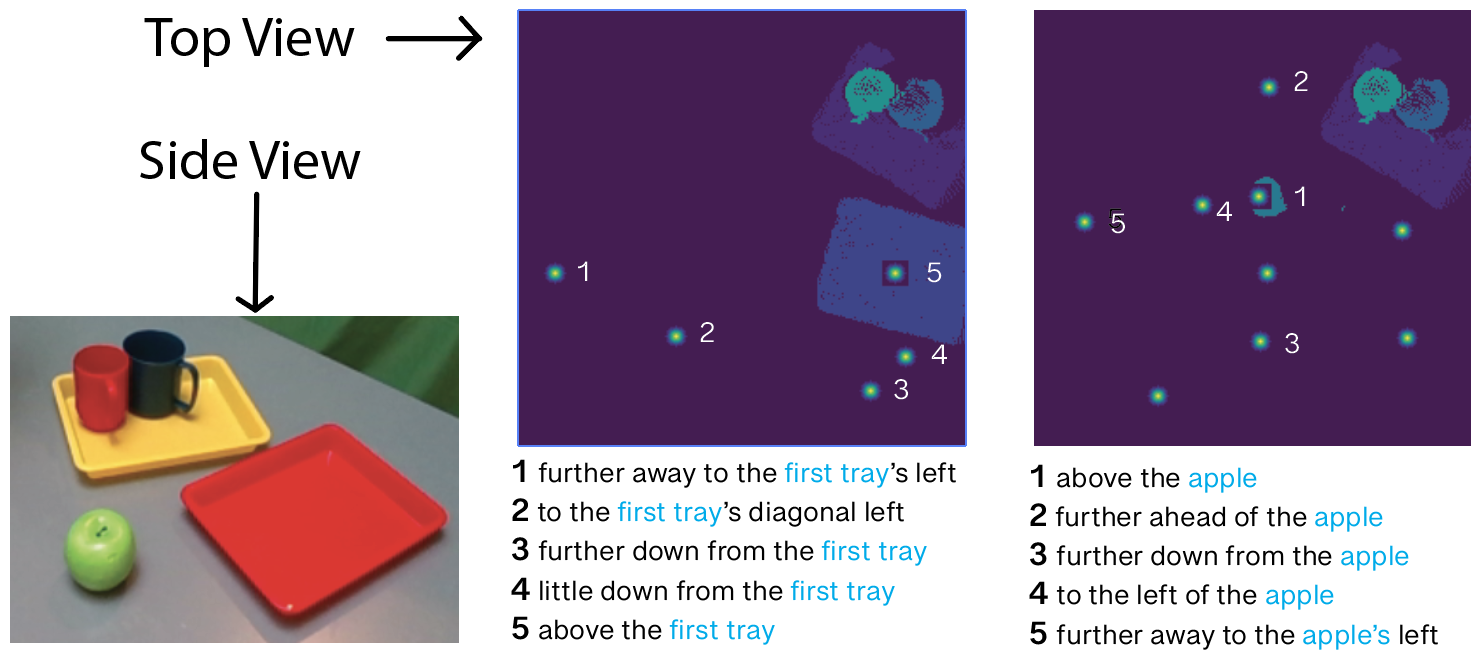}
    \caption{ Placement possibilities for an example scene. The description in \texttt{get-place-position} function argument is matched with these possibilities of the location descriptions.}
    \label{fig:placemenet_example}
\end{figure}

The place function takes in the object id to be placed, and the place position. The place position usually comes from another function called \texttt{get-place-position} which takes in the object id of the object to be placed, the object id of a reference object and a description of how the object has to be placed with respect to the reference. This description provided as an argument to \texttt{get-place-position} function is matched using BERT embeddings, with a predefined set of descriptions of place positions with respect to the reference object. These predefined descriptions are same as the relations defined between objects in scene description (Section \ref{sec:scene_desc}). Figure \ref{fig:placemenet_example} shows some of the example place positions and their descriptions with respect to the object in blue. 

\vfill
\pagebreak

\section{Python API} \label{sec:code_api}
The Python code API provided in the input prompt to the language planner is seen here:

\footnotesize
\begin{Verbatim}[commandchars=\\\{\}]
\textcolor{blue}{start_task}()
    must be called at the start of any task. It starts the robot.

\textcolor{blue}{end_task}()
    must be called when a task completes. It stops the robot.

\textcolor{blue}{get_all_object_ids}()
    returns a list of all integer object ids present in the scene; 
    Returns:
        ids: list(int)

\textcolor{blue}{get_container_id}(object_id)
    gives the id of the object that contains `object_id`
    Arguments:
        object_id: int
            id of the object that is contained in some container
    Returns:
        container_id: int or None
            the id of the container that contains `object_id`
            None is returned when the object_id is not contained in 
            any container.

\textcolor{blue}{get_objects_contained_and_over}(object_id)
    gives the ids of all the objects that lie inside or over `object_id`
    Arguments:
        object_id: int
            id of an object that contains something or over which lie 
            other objects
    Returns:
        ids: list(int)
            the ids of all the objects that lie either inside or over 
            the `object_id` an empty list is returned when nothing lies 
            over or inside.

\textcolor{blue}{find}(object_label=None, visual_description=None, place_description=None, 
     object_ids=None)
    Finds an object in the scene given atleast one of object_label, 
    visual description or place description.
    Arguments:
        object_label: str
            The name with which the object has been referred to earlier
            For example, "the second tray", "the third bowl" etc
            By default, this argument is None

        visual_description: str
            object with some visual description of what it is.
            For example, "the red mug", "the blue tray", "the checkered box"
            By default, this argument is None
            
        place_description: str
            a string that describes where the object is located. 
            For example, to find a bowl that is on the right of the 
            tray, the function call will be 
            `find("bowl that lies to the right of the tray")`, 
            or to get the mug that is contained in the second bowl, 
            the call would be 
            `find("mug that is inside the second bowl") and so on.
            By default, this argument is None

        object_ids: list(int)
            A list of `int` object ids in which the object should 
            be found, when specified.
            By default when this argument is None, all the objects 
            are considered for 
            finding the best matching 
        
        Atleast one of the first three arguments must be specified. 
        Typically, the use of the first and 
        the second argument is enough but third can be used whenver needed.
            
    Returns: int
        object_id, an integer representing the object that best 
        matched with the description

\textcolor{blue}{get_location}(object_id)
    gives the location of the object `object_id`
    Arguments:
        object_id: int
            Id of the object
    Returns:
        position: 3D array
            the location of the object

\textcolor{blue}{pick}(object_id)
    Picks up an object that `object_id` represents. A `place` needs 
    to occur before another call to pick, i.e. two picks cannot 
    occur one after the other
    Arguments:
        object_id: int
            Id of the object to pick
    Returns: None
    
\textcolor{blue}{get_place_position}(object_id, reference_object_id, place_description)
    Finds the position to place the object `object_id` at a location 
    described by `place_description` with respect to the 
    `reference_object_id`. 
    Arguments:
        object_id: int
            Id of the object to place
        reference_object_id: int
            id of the object relevant for placing the object_id
        place_description: str
            a string that describes where with respect to the 
            reference_object_id the object_id should be placed.

    Returns: 3D array
        the [x, y, z] value for the place location is returned.

    For example, 
    to place a mug to the left of a bowl, the following function 
    call should be used
        get_place_position(mug_id, bowl_id, "to the left")
    to place a mug into a bowl:
        get_place_position(mug_id, bowl_id, "inside")
    to place a mug above a box:
        get_place_position(mug_id, box_id, "above")

\textcolor{blue}{place}(object_id, position)
    Moves to the position and places the object `object_id`, at the location 
    given by `position` with the same orientation the object is currently in. 
    The robot will open the gripper and drop the item at the position.
    Arguments:
        object_id: int
            Id of the object to place
        position: 3D array
            the place location
    Returns: None

\textcolor{blue}{learn_skill}(skill_name)
    Adds a new category-level skill to the current list of skills.
    Arguments:
        skill_name: str
            a short name for the skill to learn, must be a string that can 
            represent a function name (only alphabets and underscore can be used).  
            highly-recommended that the object labels are included in the skill name.
    Returns:
        skill_function: method
            a function that takes as input an object_id and 
            performs the skill on the object represented by the object_id
            another relevant object_id can be passed optionally. In particular,
            the returned function takes in arguments: object_id_1 and object_id_2:
            object_id_1: int
                Id of the object to act upon
            object_id_2: int (optional)
                Id of the object to place/interact relative to if relevant
        skill_docstring: str
            a string that describes how to use the new skill in words, including relevant 
            inputs and outputs, 
            along with any information on appropriate situations to use the skill,
            and misleading/confusing scenarios where it might make sense to use 
            the skill but where a different skill should actually be used.  
            This docstring should be printed out in the console so that the user can 
            copy it and paste it into the skill API for use
            on subsequent runs (since the docstring will provide helpful context for 
            how to appropriately use the skill in the future). 

    For example:
        # Example 1:
        drawer_id = 2
        open_drawer, open_drawer_doc = learn_skill("open_drawer")
        print(open_drawer_doc)
        [Out:] 
            Grasps the handle of the drawer and executes a linear motion 
            in the direction away from the drawer, so that the drawer opens.
            Arguments: 
                object_id: int
                    ID of the drawer to open
            Returns: None
        # opens the drawer represented by drawer_id
        open_drawer(drawer_id) 

        # Example 2:
        stick_id = 3
        ball_id = 4
        hit_ball_with_stick, hit_ball_with_stick_doc = learn_skill("hit_ball_with_stick")
        print(hit_ball_with_stick_doc)
        [Out:] 
            Hits a ball with a stick that is currently held in the gripper. 
            Hitting is performed by moving the stick to an offset position away 
            from the ball (depending on the current position of the ball obtained 
            by the ball point cloud and the size of the stick, obtained 
            by the stick point cloud) and then executing a fast gripper + 
            wrist motion to move the stick and hit the ball. Can only be executed 
            after executing some kind of "pick" for the stick, so that the stick 
            is in the grasp.
            Args: 
                object_id_1: int
                    ID of the stick to use for hitting
                object_id_2: int
                    ID of the ball that should be hit with the stick
            Returns: None
        # hits the ball represented by ball_id with the stick represented by stick_id
        hit_ball_with_stick(stick_id, ball_id)

\end{Verbatim}

\vfill
\pagebreak

\normalsize
\section{Experiments}
\subsection{Learning a new skill}
\begin{figure}[H] 
    \centering
    \includegraphics[width=\linewidth]{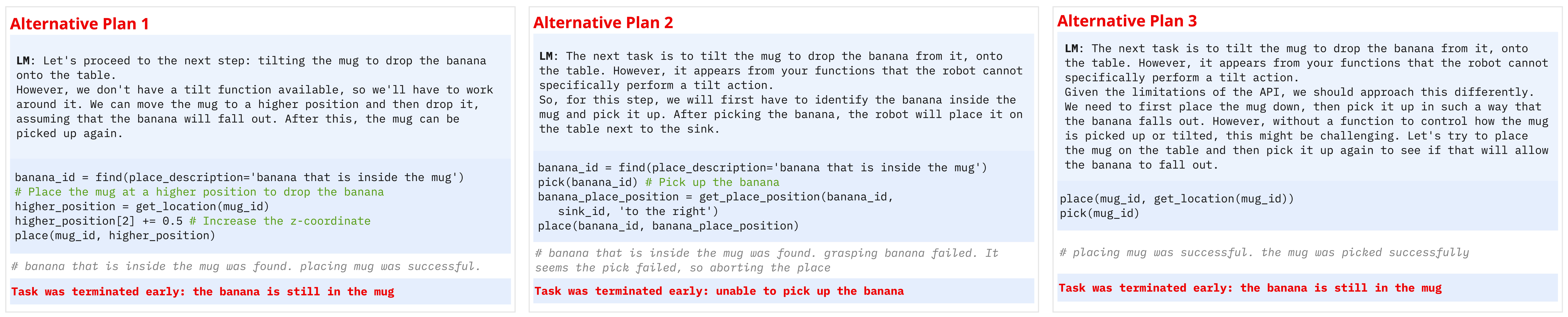}
    \caption{ Alternate suggestions from the LM to remove a banana from a mug, when learning a new skill is not an option. }
    \label{fig:failed_tilt}
\end{figure}

Additional planner outputs for the ``Empty the sink out" tasks in the previously described scene are provided in Figure \ref{fig:failed_tilt} to demonstrate the alternative nonsensical plans when the LM has to conform to the static skill API. 

\begin{figure}[H] 
    \centering
    \includegraphics[width=0.75\linewidth]{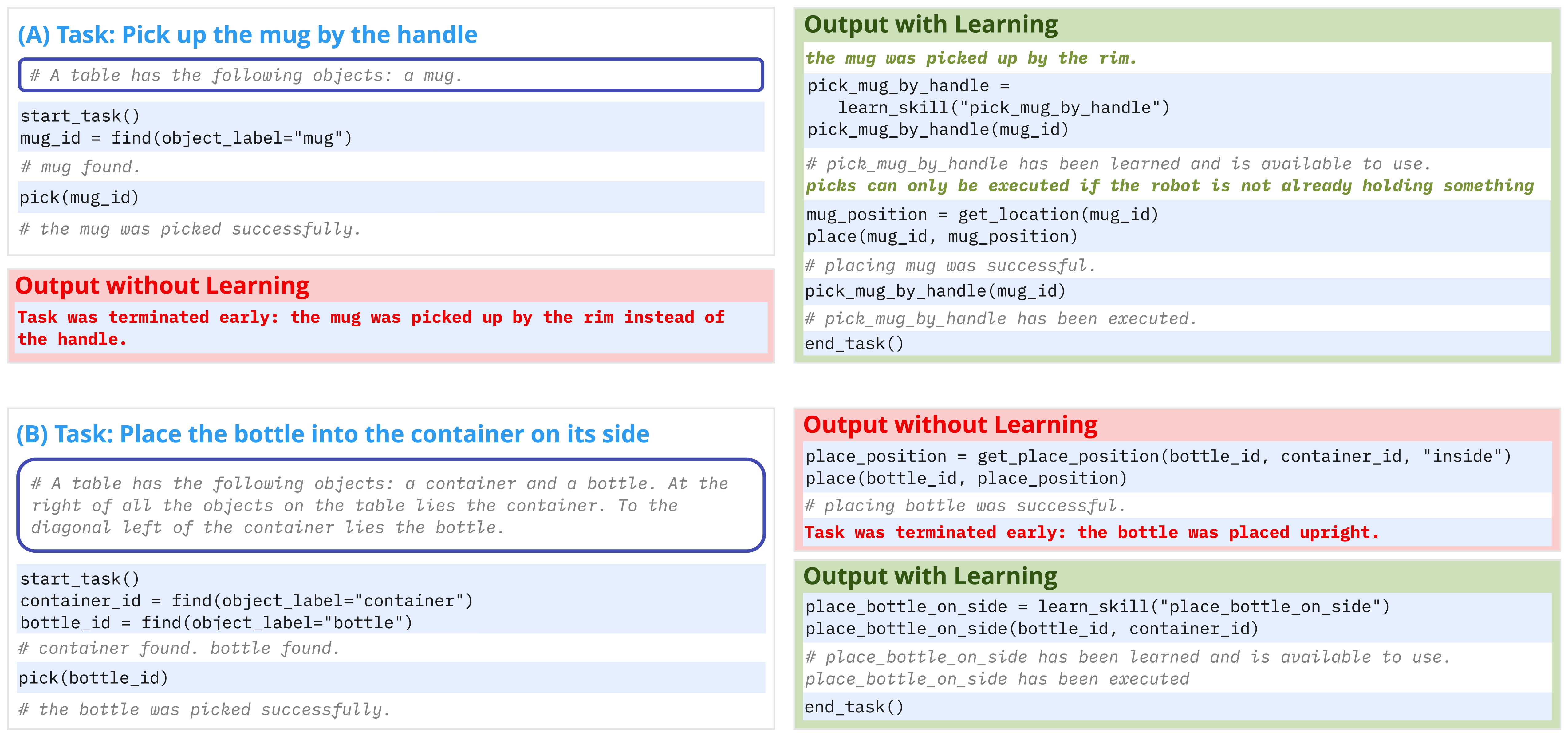}
    \caption{ Planner outputs for additional tasks using baseline compared to our system.}
    \label{fig:more_learning}
\end{figure}

With learning capabilities, it generates programs that accomplish the more difficult tasks by acquiring new policies to execute a handle grasp and a sideways reorientation placement seen in Figure \ref{fig:more_learning}.

\subsection{Scene Description Ablations}
\begin{figure}[H]
    \centering
    \setlength{\tabcolsep}{0.3pt}
    \begin{tabular}{cc}
    \begin{subfigure}[t]{0.425\linewidth}
        \centering 
        \includegraphics[width=\linewidth]{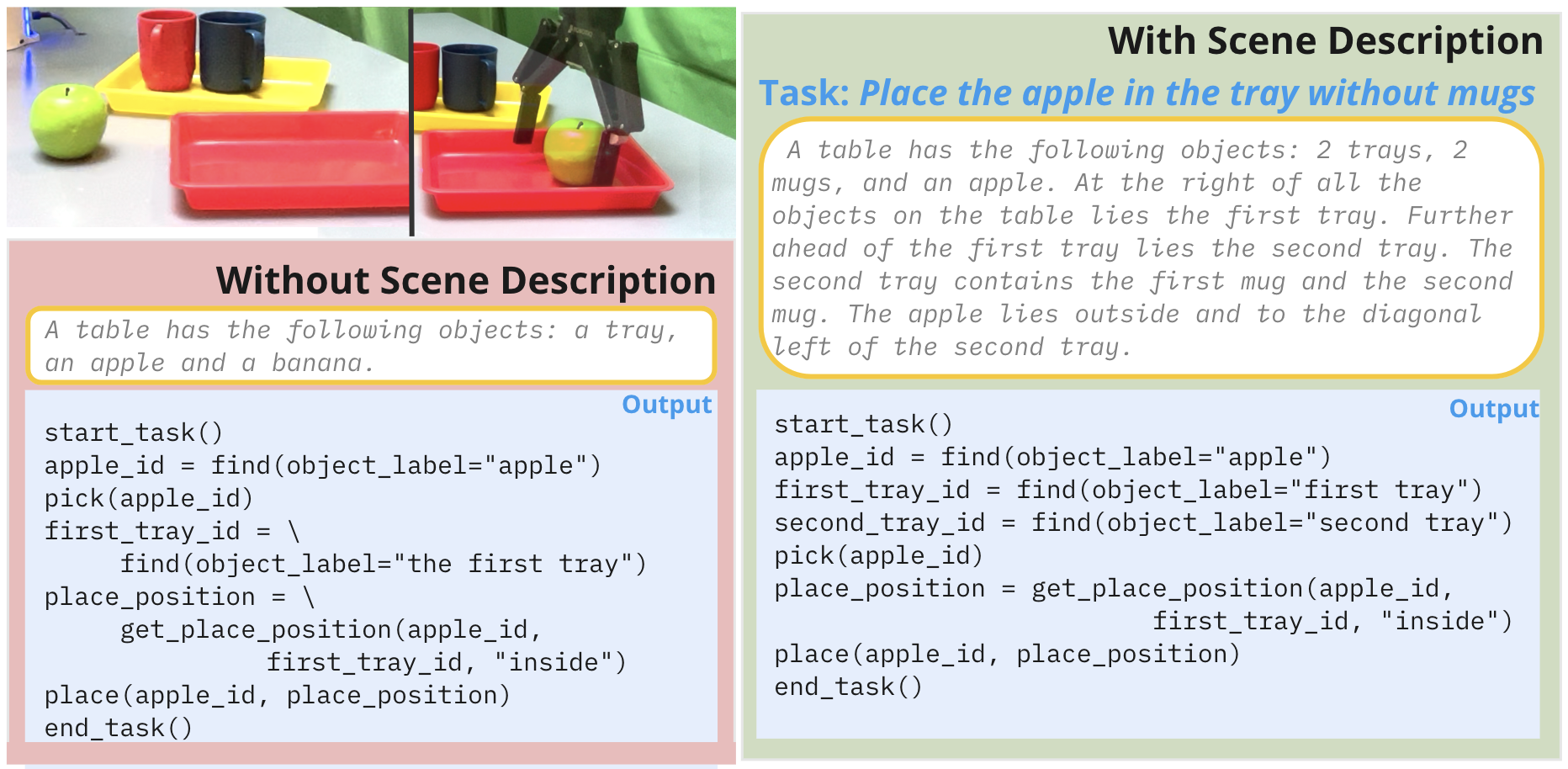}
        \caption{ Planner output for tasks in scene description category with indirect reference to an object.}
        \label{fig:apple}
    \end{subfigure} \hspace{30pt} \hfill
    \begin{subfigure}[t]{0.5\linewidth}
      \centering
        \includegraphics[width=\linewidth]{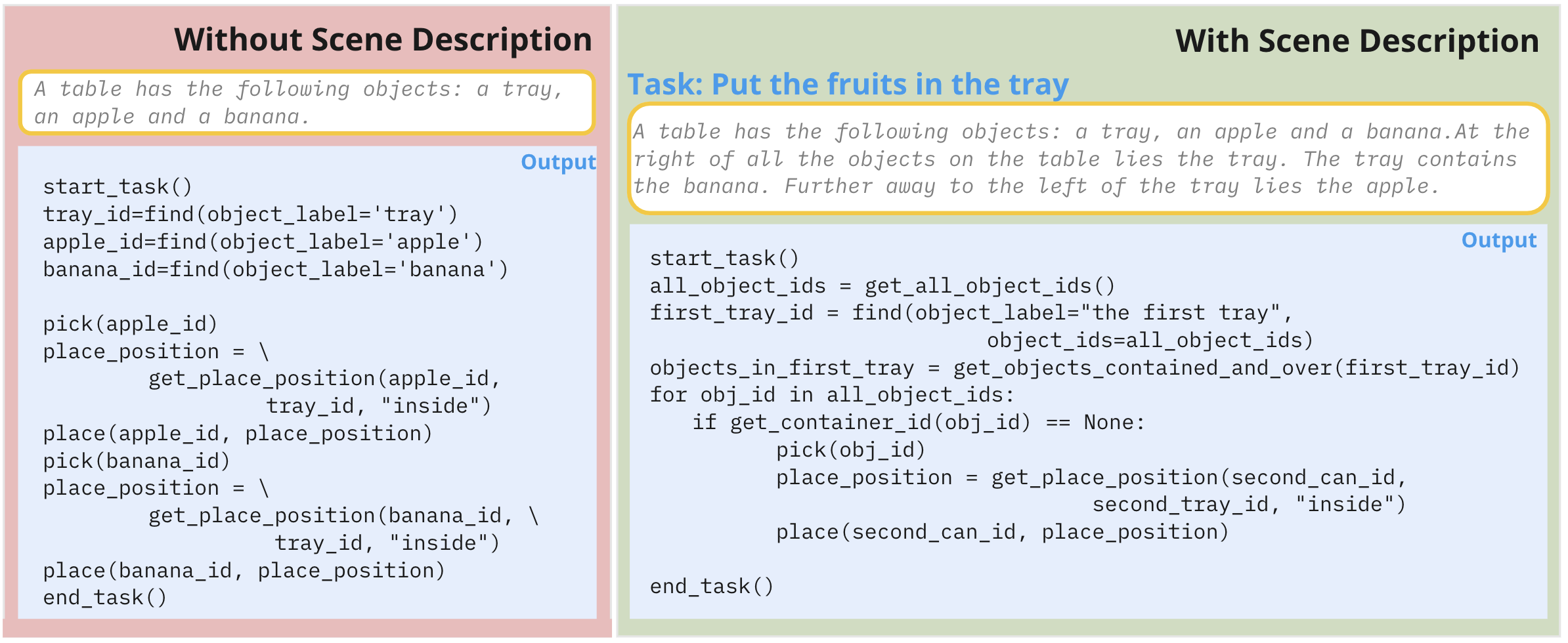}
        \caption{ Planner output for scene description category involving detecting task progress.}
        \label{fig:banana}

    \end{subfigure}
    \end{tabular}

\end{figure}

Plans for experiments in the scene description category are also shown in Figure \ref{fig:apple} and Figure \ref{fig:banana}.

\vfill
\pagebreak

\section{Limitations}

\textbf{Scene Descriptions}
The complexity of the textual description is an ongoing research problem, and it is difficult to capture every visual feature needed to plan optimally. Camera sensor noise, clutter, and failure to consolidate multiple views of the same object instance are all factors that may cause perception to fail. The perception module may misidentify some objects, while noise or clutter can cause spatial relationship heuristics to fail, thus generating inaccurate scene description that results in incorrect plans. 

\textbf{Function Usage Errors}
Functions that take in open-vocabulary text as parameters are often used incorrect or insufficiently by the LM. The system may then misidentify placements or objects due to incorrect correspondence of the language embeddings. In addition, occasionally the planner will request a new skill with some assumption about its parameters that are unknown until usage. Therefore the user may misinterpret the request and teach it the wrong skill. Finally the planner may request an entire sequence of actions as a new skill which goes against the purpose of the function. An improved API with better function descriptions and more examples may resolve the issue.

\textbf{Skill Primitives Failure}
While executing pick and place primitives, the generated robotic arm motion plans may be infeasible due to joint limits, or failure in inverse kinematics. This lowers the success rate of the primitive skill functions. Another failure case in placements is when an object has to be placed inside another object, but due to noise in execution the object topples out of the container, thus failing the task.

\textbf{Feedback Automation}
Our generated textual scene description and execution feedback aims to mitigate the amount of user feedback needed to form a closed-loop system. We fallback on user feedback to describe object states of actions that go beyond successful control execution, which is difficult to automate (see \ref{subsec:learning_feedback}). To build a self-sufficient embodied agent, a more sophisticated verification pipeline  with automatic environment and object state feedback is required. Some systems use multimodal inputs similar to \cite{driess2023palme} as feedback.

\textbf{GPT-3.5 vs GPT-4}
The code generation capabilities of GPT-4 far exceeds that of GPT-3.5, but we have a limited number of messages we are able to send, so it is difficult to collect a large amount of experiment data for that reason.

\section{Tasks for LLM-only \learnskill{} and skill re-use evaluation}
In the subsections below, we include a brief description of each task used to evaluate the LLM on its own in its ability to request new skills, avoid requesting unnecessary skills, and perform tasks with an expanded skill library after a newly-learned skill is added. Each of these tasks was manually curated by hand and did not reflect any real-world environment, nor were the plans meant to be executable on the robot. For brevity, we include the full prompt (including the manually designed scene description) of one of the tasks for each section, and only include a concise description of the task to be solved for the rest. 
\paragraph{Ask for a new skill when needed --}
We considered the tasks listed below for evaluating how frequently GPT-4 correctly requests a new skill to be learned using the \learnskill{} function:
\begin{itemize}[leftmargin=*]
    \item mix the ingredients (a bowl and a spoon, bowl is filled)
    \begin{verbatim}
A table has the following objects: a bowl and a spoon. 
At the right of all the objects on the table lies the bowl. 
The bowl is filled. To the left of the bowl lies the spoon. 
If you are commanding a robot, tell me in words the steps 
to mix the ingredients? 
    \end{verbatim}
    \item empty the mug (mug contains water)
    \item hang the mug on the mug rack
    \item place the mug upside down
    \item bottle in the container horizontally
    \item put the bowl on its side on the drying rack
    \item scrub the bowl
    \item open the drawer
    \item slide the tray into the oven
    \item push the button
    \item turn on the light
\end{itemize}

\paragraph{Don’t ask for a new skill when \emph{not} needed --}
Contrasting the above scenario, the tasks listed below are used to evaluated how frequently the planner \emph{ignores} the \learnskill{} function and directly attempts to complete the task with the base set of primitive skills that are available. The tasks are constructed such that they can be achieved using the base set of skills: 
\begin{itemize}[leftmargin=*]
    \item build a block tower in rainbow color order
    \begin{verbatim}
A table has the following objects: a red block, blue block, 
and green block. At the right of all the objects on the table 
lies the blue block. To the left of the blue block lies the red
block. To the left of the second bowl lies the green block. 
If you are commanding a robot, tell me in words the steps to 
build a block tower in rainbow color order? 
    \end{verbatim}
    \item stack the bowls
    \item put the dishes away
    \item get me a cup I can pour water into
    \item empty the bowl
    \item put the dishes in the sink
\end{itemize}

\paragraph{When new skills have been added, how often the agent succeeds at a new task which may or may not require the new skill --}
Besides evaluating the LLM's ability to correctly use the base skill set and request new abilities with \learnskill{}, we also evaluated the planner's success when prompted with new tasks after newly-learned skills have been added to the API. These tasks have been constructed to sometimes require the use of the newly-learned skill and other times specifically require \emph{not} using the newly learned skill. The tasks we use for this evaluation are listed below: \\

\textbf{After learning and adding \texttt{pick\_mug\_by\_handle} to API:}
\begin{itemize}[leftmargin=*]
    \item pick up the mug (handle of the mug is broken)
    \begin{verbatim}
A table has the following objects: a mug. At the right of all the
objects on the table lies the mug. The handle of the mug is broken. 
If you are commanding a robot, tell me in the words the steps to 
pick up the mug? 
    \end{verbatim}
    \item empty the mug
    \item pick up the mug (box is blocking access to the handle)
    \item bring the cup of coffee to the living room
    \item pick up the mug (handle of the mug is covered with a dirty, sticky substance)
    \item fill the mug with water
\end{itemize}
\textbf{After learning and adding \texttt{side\_pick\_bottle} to API:}
\begin{itemize}[leftmargin=*]
    \item fetch the bottle and put it on the table (bottle is on a bottom shelf, cannot be approached from the top)
    \begin{verbatim}
A table has the following objects: a bottle. At the right of all 
the objects on the table lies the bottle. Due to the position on 
the table and the low height of the table, the bottle cannot be 
reached from the side (but could be reached from above). If you 
are commanding a robot, tell me in the words the steps to fetch 
the bottle? 
    \end{verbatim}
    \item fetch the bottle (bottle cannot be reached from the side, but could be reached from above)
    \item fetch the bottle and put it on the table (bottle inside a box, box is sideways, top of the box is open)
    \item fetch the bottle and put it on the table (bottle is on a bottom shelf)
\end{itemize}
\textbf{After learning and adding \texttt{place\_book\_horizontally} and \texttt{place\_book\_vertically} to API:}
\begin{itemize}[leftmargin=*]
    \item put away the book (bookshelf with single book, book doesn’t fit vertically)
    \begin{verbatim}
A table has the following objects: a book and a bookshelf. At the
right of all the objects on the table lies the bookshelf. Next to
the bookshelf lies the book. The book doesn’t fit in the shelf
vertically. If you are commanding a robot, tell me in the words the
steps to put away the book?
    \end{verbatim}
    \item put away the book (bookshelf with single book, bookshelf contains many vertically aligned books, with space for one more)
    \item put away the book (bookshelf with single book, book LxWxH and bookshelf LxWxH dimensions provided)
    \item put away all of the books (bookshelf and 10 books total, only three can fit when horizontal)
\end{itemize}

\twocolumn

\end{document}